\documentclass[runningheads,a4paper]{llncs}

\usepackage{amssymb}
\usepackage{graphicx}
\usepackage{multirow}
\usepackage{longtable}
\usepackage{rotating}
\usepackage{color}
\graphicspath{{figures/}}

\usepackage{url}
\urldef{\mailsa}\path||
\urldef{\mailsb}\path||
\urldef{\mailsc}\path||    
\newcommand{\keywords}[1]{\par\addvspace\baselineskip
\noindent\keywordname\enspace\ignorespaces#1}

\begin{document}

\mainmatter  

\title{Multi-focus Image Fusion using dictionary learning and Low-Rank Representation}

\titlerunning{Multi-focus Image Fusion method: dictionary learning and LRR}

%
%
\author{Hui Li\and Xiao-Jun Wu%
\thanks{xiaojun\_wu\_jnu@163.com}%
}
\authorrunning{Multi-focus Image Fusion method based on dictionary learning and LRR}

\institute{School of Internet of Things Engineering,\\ Jiangnan University, Wuxi 214122, China.
\mailsa\\
\mailsb\\
\mailsc\\
\url{}}

%
%

\toctitle{Multi-focus Image Fusion method}
\tocauthor{School of Internet of Things Engineering, Jiangnan University}
\maketitle

\begin{abstract}
Among the representation learning, the low-rank representation (LRR) is one of the hot research topics in many fields, especially in image processing and pattern recognition. Although LRR can capture the global structure, the ability of local structure preservation is limited because LRR lacks dictionary learning. In this paper, we propose a novel multi-focus image fusion method based on dictionary learning and LRR to get a better performance in both global and local structure. Firstly, the source images are divided into several patches by sliding window technique. Then, the patches are classified according to the Histogram of Oriented Gradient (HOG) features. And the sub-dictionaries of each class are learned by K-singular value decomposition (K-SVD) algorithm. Secondly, a global dictionary is constructed by combining these sub-dictionaries. Then, we use the global dictionary in LRR to obtain the LRR coefficients vector for each patch. Finally, the $l_1-norm$ and choose-max fuse strategy for each coefficients vector is adopted to reconstruct fused image from the fused LRR coefficients and the global dictionary. Experimental results demonstrate that the proposed method can obtain state-of-the-art performance in both qualitative and quantitative evaluations compared with serval classical methods and novel methods. The code is available at \url{https://github.com/hli1221/imagefusion_dllrr}.
\end{abstract}

\section{Introduction}

Image fusion is an important technique in image processing community. The main purpose of image fusion is to generate a fused image by integrating complementary information from multiple source images of the same scene \cite{Shutao2017}. And the multi-focus image fusion is a branch of image fusion. For image with deep depth field which is very common, it usually contains clear(focus) and blurry(defocus) parts. With the help of multi-focus image fusion technique, the image of the same scene can be combined into a single all-in-focus image. In recent years, the image fusion technique has become an active task in image processing community, and it has been used in many fields, such as medical imaging, remote sensing and computer vision.

	Various algorithms for multi-focus image fusion have been developed over the past decades. Early image fusion methods mainly focus on non-representation learning-based methods. And the multi-scale transforms are the most commonly methods, the classical methods include discrete wavelet transform(DWT) \cite{DWT2005}, and other transform domain methods \cite{contourlet2010,shearlet2014,quaternion wavelet2013}. Due to the classical transform methods has not enough detail preservation ability, Luo X et al. \cite{nonshearlet2017} proposed contextual statistical similarity and nonsubsampled shearlet transform based fusion method. In reference \cite{mwgfm2014}, Zhou Z et al. proposed a novel fusion method base on multi-scale weighted gradient. And this method has better detail preservation ability than above methods. Recently, the morphology which is also a non-representation learning technique was applied to image fusion. Yu Zhang et al. \cite{morphological2017} proposed a morphological gradient based fusion method. The focus boundary region, focus region and defocus region are extracted by different morphological gradient operator, respectively. Finally, the fused image is obtained by using an appropriate fusion strategy.
	
	As we all know, the most common methods of representation learning are work in sparse domain. Yin H et al. \cite{novelsr2016} proposed a novel sparse representation-based fusion method which use the source images as train data to obtain a joint dictionary, then use a maximum weighted multi-norm fusion rule to reconstruct fused image from the dictionary and sparse coefficients. But the detail preservation ability of this method is not well, so Zong J J et al. \cite{hogsr2017} proposed a fusion method based on sparse representation of classified images patches, which used the Histogram of Oriented Gradient(HOG) features to classify the image patches and learned serval sub-dictionaries. Then this method use the $l_1-norm$ and choose-max strategy to reconstruct fused image. And the fusion method based on the joint sparse representation and saliency detection was proposed by Liu C H et al. \cite{jointsparse2017} and obtained great fusion performance in infrared and visible image fusion. Besides the above methods, the convolutional sparse representation \cite{convolutional2016} and cosparse analysis operator \cite{cosparse2017} were also introduced to image fusion task. Recently, a fusion method based on convolutional neural network (CNN) also proposed by Yu Liu et al. \cite{cnn2017}.

	Although the sparse representation based fusion method has many advantages, the ability of capture global structure is limited. On the contrary, the low-rank representation (LRR) \cite{lrr2010} could capture the global structure of data, but could not preservation the local structure. So in this paper, we proposed a novel multi-focus image fusion method based on dictionary learning and LRR to get a better performance in both global and local structure and this method will be introduced in next sections. The experimental results demonstrate that the proposed method can obtain very good fusion performance.

	The rest of this paper is organized as follows. In Section 2, we give a brief introduction to related work include LRR theory and dictionary learning. In Section 3, the proposed dictionary learning and LRR-based image fusion method is presented in detail. The experimental results and discussions are provided in Section 4. Finally, Section 5 concludes the paper.

\section{Related Work}

The dictionary learning and LRR theory are two major parts in our fusion method. The LRR theory insures that we could capture the global structure of input data. And the dictionary learning processing insures the local structure information could be captured accurately.

	The K-singular value algorithm (K-SVD) \cite{ksvd2006} is a standard unsupervised dictionary learning algorithm which is widely investigated in many fields. In this paper, the K-SVD algorithm is used to learn the dictionary. 
	
The LRR theory is an important representation learning method. In reference \cite{lrr2010}, authors apply self-expression model to avoid training a dictionary and the LRR problem will be solved by the following optimization problem,
\begin{eqnarray}\label{equ:lrr}
  	\min_{Z,E}||Z||_*+\lambda||E||_{2,1} \\
    s.t.,X=XZ+E \nonumber
\end{eqnarray}
where $X$ denotes the observed data matrix, $||\cdot||_*$ denotes the nuclear norm which is the sum of the singular values of matrix. $||E||_{2,1}=\sum_{j=1}^{n}\sqrt{\sum_{i=1}^{n}[E]_{ij}^2}$ is called as $l_{2,1}-norm$, $\lambda>0$ is the balance coefficient. Eq.\ref{equ:lrr} is solved by the inexact Augmented Lagrange Multiplier (ALM). Finally, the LRR coefficients matrix $Z$ for $X$ is obtained by Eq.\ref{equ:lrr}.

\section{The Proposed Image Fusion Method}

In this section, dictionary learning and low-rank representation based multi-focus image fusion method is presented in detail. The framework of our method is shown in Fig.\ref{fig:dictionarylearning} and Fig.\ref{fig:proposedmethod}. The Fig.\ref{fig:dictionarylearning} is the diagram of dictionary learning processing. And Fig.\ref{fig:proposedmethod} is the diagram of our fusion method.

\subsection{Dictionary learning}

\begin{figure}[ht]
\centering
\includegraphics[width=\linewidth]{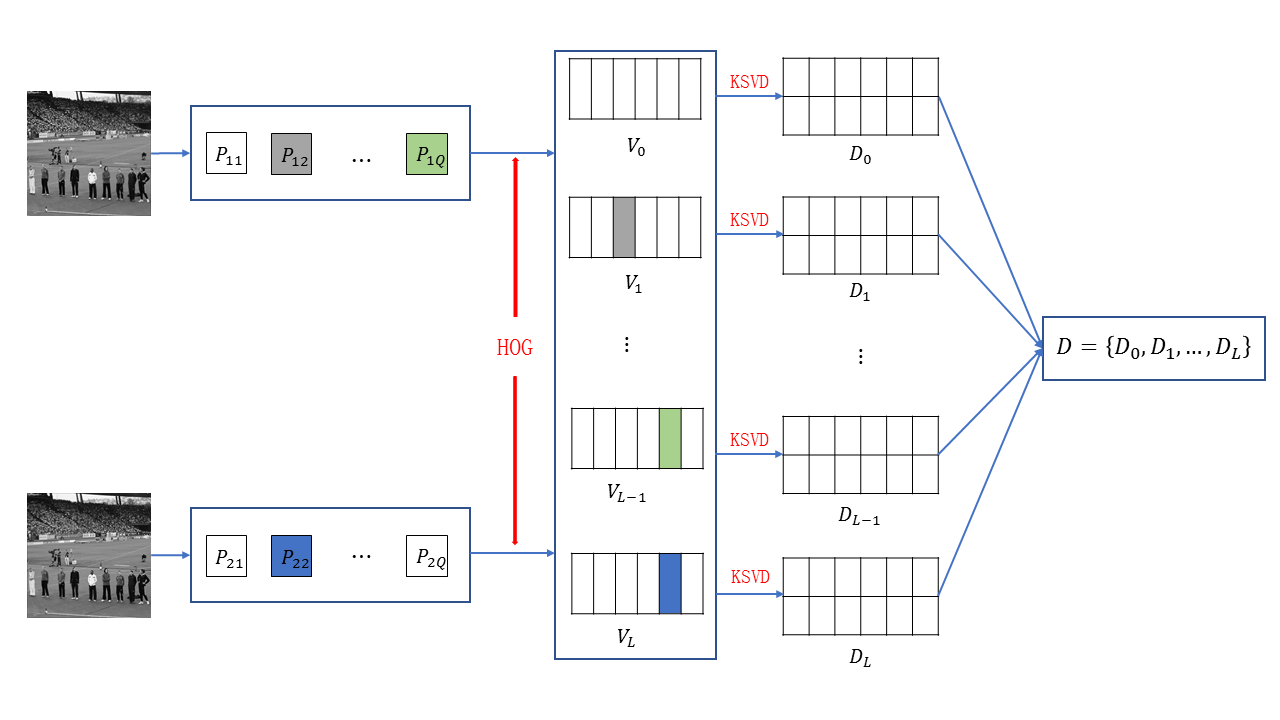}
\caption{The framework of sub-dictionary learning and obtain joint dictionary.}
\label{fig:dictionarylearning}
\end{figure}

In this section, we will introduce the dictionary learning method in our fusion method. As shown in Fig.\ref{fig:dictionarylearning}, for the case of two source images which are denoted as $I_A$ and $I_B$ respectively. Then image patches are obtained by using sliding window technique for each source image. Assume that the size of $I_A$ and $I_B$ is $N$$\times$$M$, the windows size is $n$$\times$$n$, and the step is $s$, so the patches number for each source image is $Q=\left(\lfloor (\frac{N-n}{s}) \rfloor+1 \right) $$\times$$ \left(\lfloor (\frac{M-n}{s}) \rfloor+1 \right)$. If the source images number is two, there are altogether $2Q$ image patches.

Then the Histogram of Oriented Gradient(HOG) features\cite{hog2009} are used to classify these patches. In the procedure of extract HOG features, suppose that there are $L$ orientation bins $\{\theta_1,\theta_2,\cdots,\theta_L,\}$. And $G_i (\theta_j )(j=1,2,\cdots,L)$ denotes the gradient value in the $j-th$ orientation bin for the $i-th$ patch $P_i$. Define the $J_i$ as the class of patch $P_i$. The rules for classification are as follows,

\begin{eqnarray}\label{equ:hogclass}
  	J_i=\left\{\begin{array}{ll}
		0 & \textrm{ $\frac{G_imax}{\sum_{j=1}^{L}G_i(\theta_j)}<T$ } \\
        J & \textrm{ otherwise }
	\end{array}\right.
\end{eqnarray}

\noindent where $G_imax=\max⁡{\{G_i (\theta_i)\}}$ is the maximum oriented gradient value for image patch $P_i$. And the index of $G_imax$ is $J=arg\max_{J}{⁡\{G_i(\theta_i)\}}$, which represents the most dominant orientation of patch  $P_i$. $T$ is a threshold to determine whether the patch has certain dominant orientation or not. $J_i=0$ means the patch does not have any dominant orientation, which implies the patch is irregular. Otherwise, the patch belongs to the category $J$.

After the classification, all the classified patches are reconstructed into lexicographic ordering vectors, which constitute corresponding matrix $V_j (j=0,1,\cdots,L)$. For each matrix $V_c$, the corresponding dictionary $D_j$ is obtained by K-SVD algorithm. The sub-dictionaries $D_j (j=0,1,…,L)$ are obtained, respectively. Then a global dictionary $D$ is constructed by combining these sub-dictionaries, as shown in Fig.1. This global dictionary $D$ will be used in the procedure of image fusion and as a dictionary for LRR.

\subsection{Proposed Fusion Method}

In the procedure of image fusion, firstly, each source images are divided into $Q$ image patches, as discussed in section 3.1. Then all the patches are transformed into vectors via lexicographic ordering, the image patches matrix $VI_A$ is constituted by combine these vectors. And the image patches matrix $VI_B$ is obtained by the same operation for source image $I_B$.

\begin{figure}[ht]
\centering
\includegraphics[width=\linewidth]{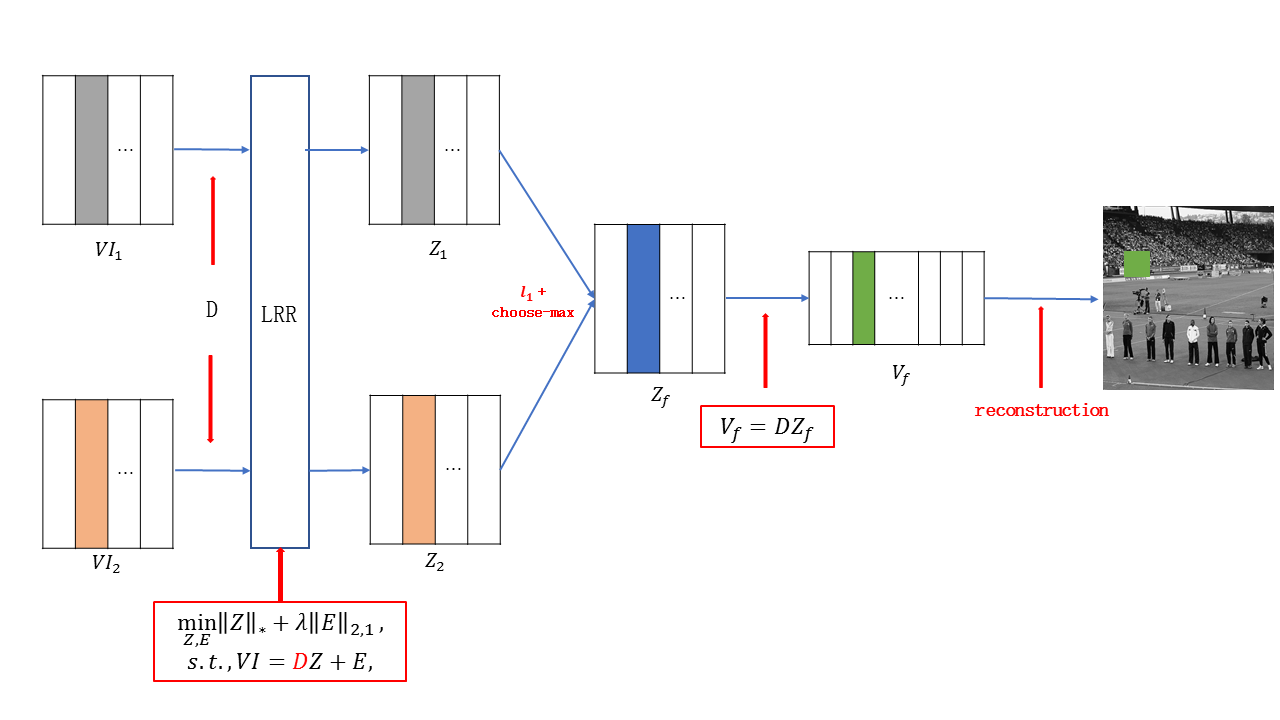}
\caption{The framework of proposed fusion method.}
\label{fig:proposedmethod}
\end{figure}

As shown in Fig.\ref{fig:proposedmethod}, the LRR coefficients matrices $Z_A$ and $Z_B$ are calculated by matrices $VI_A$, $VI_B$ and Eq.\ref{equ:dlrr}.
\begin{eqnarray}\label{equ:dlrr}
  	\min_{Z_C,E}||Z_C||_*+\lambda||E||_{2,1} \\
    s.t.,VI_C=DZ_C+E, C\in{\{A,B\}} \nonumber
\end{eqnarray}

\noindent where $VI_C, C\in{\{A,B\}}$ donates the coefficients matrix which obtained from $I_A$ or $I_B$. The dictionary $D$ is a global dictionary obtained by combined sub-dictionaries as discussed in section 3.1. $Z_C$ is the LRR coefficients matrix for $VI_C$. We use the inexact ALM to solve the problem \ref{equ:dlrr}. 

Let $Z_Ci$ denotes the $i-th$ column of $Z_C$, where $i=\{1,2\cdots,Q\}$. Then $Z_Ci$ is the LRR coefficients vector for corresponding image patch $P_Ci$. Then we use $l_1-norm$ and choose-max strategy to fuse the corresponding LRR coefficients vector, the fused LRR coefficients vector is obtained by Eq.\ref{equ:choosecoe},
\begin{eqnarray}\label{equ:choosecoe}
  	Z_fi=\left\{\begin{array}{ll}
		Z_Ai & \textrm{ $||Z_Ai||_1>||Z_Bi||_1$ } \\
        Z_Bi & \textrm{ otherwise }
	\end{array}\right.
\end{eqnarray}

\noindent Using Eq.\ref{equ:choosepatch}, we will get fused LRR coefficients matrix$Z_f$.

Finally, the fused image patches matrix $V_f$ is obtained by Eq.\ref{equ:choosepatch}. In this formula, $D$ is the global dictionary which is obtained by combined sub-dictionaries. And $Z_f$ is the fused coefficients matrix which obtained by Eq.\ref{equ:choosecoe}.
\begin{eqnarray}\label{equ:choosepatch}
  	V_f=DZ_f
\end{eqnarray}

Let $V_fi$ denotes the $i-th$ column of $V_f$, where $i=\{1,2\cdots,Q\}$. Reshape the vector $V_fi$ as a $n$$\times$$n$ patch and take the patch back to its original corresponding position. And a simple averaging operation is applied to all overlapping patches to form the fused image $I_F$.

\subsection{Summary of the Proposed Fusion Method}

To summarize all the previous analyses, the procedure of the dictionary learning and Low-Rank Representation based multi-focus fusion method is described as follows:

	1) For each source image, it is divided into $Q$ image patches by sliding window technique.
	
	2) A global dictionary $D$ is obtained by combined sub-dictionaries.
	
		\ a. These image patches are classified based on HOG features in which each dominant orientation corresponds to a class.
		
		\ b. For each class, the sub-dictionary is calculated by K-SVD algorithm.
		
		\ c. These sub-dictionaries are united as a global dictionary.
		
	3) All the patches which from a source image are transformed into vectors via lexicographic ordering, and constitute the image patches matrix $VI_A$ and $VI_B$.
	
	4) The LRR coefficients matrices $Z_C$ for each source image are obtained by LRR and the learned global dictionary.
	
	5) The fused LRR coefficients matrix $Z_f$ is obtained. For each corresponding column of LRR coefficients matrix, we use $l_1-norm$ and choose-max strategy to fuse the coefficients vector.
	
	6) The fused image patches matrix $V_f$ are obtained by Eq.\ref{equ:choosepatch}.
	
	7) Averaging operation is applied to all overlapping patches to reconstruct the fused image $I_F$ from $V_f$.

\section{Experiment}

	This section firstly presents the detailed experimental settings and introduce the source images which we choose in our experimental. Then the fused result of proposed method and other method are analyzed.

\subsection{Experimental settings}

\begin{figure}[!ht]
\centering
\includegraphics[width=0.27\linewidth]{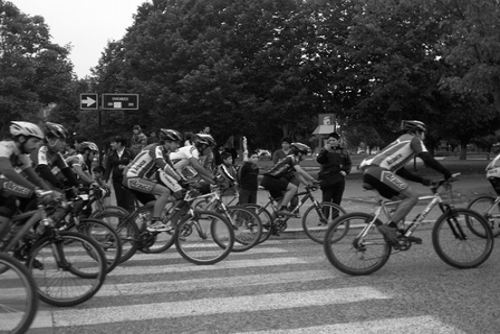}
\includegraphics[width=0.27\linewidth]{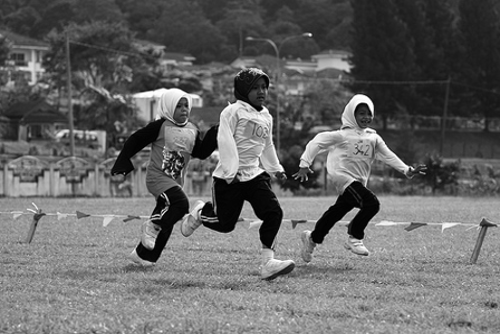}
\includegraphics[width=0.27\linewidth]{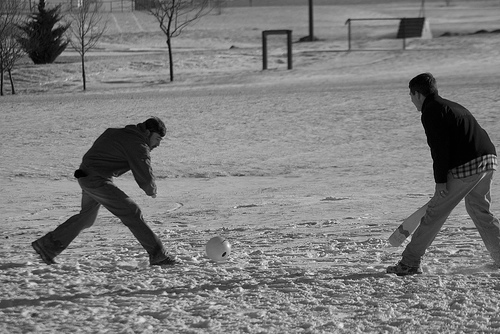} \\
\includegraphics[width=0.27\linewidth]{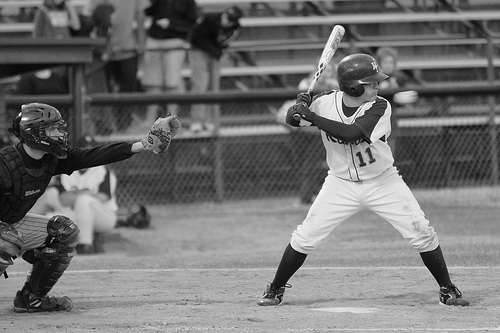}
\includegraphics[width=0.27\linewidth]{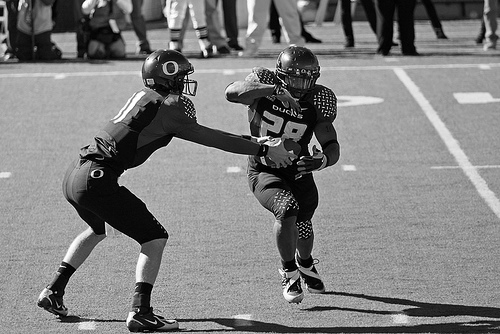}
\includegraphics[width=0.27\linewidth]{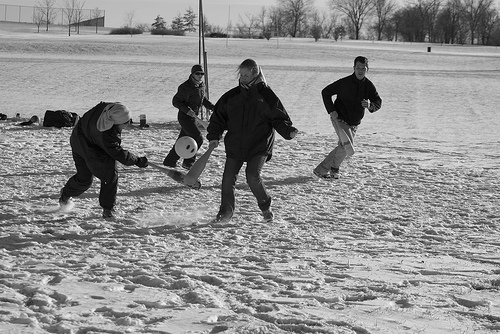} \\
\caption{ Six original images from the twenty original images.}
\label{fig:example6}
\end{figure}

\begin{figure}[!ht]
\centering
\includegraphics[width=0.4\linewidth]{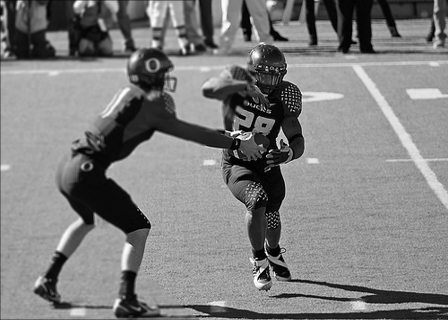}
\includegraphics[width=0.4\linewidth]{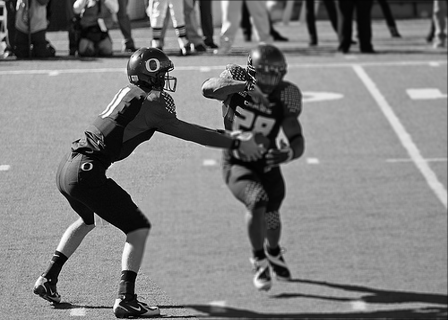} \\

\includegraphics[width=0.4\linewidth]{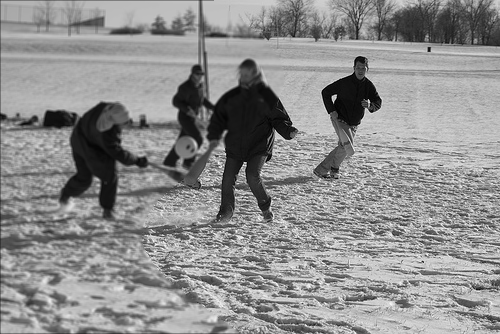}
\includegraphics[width=0.4\linewidth]{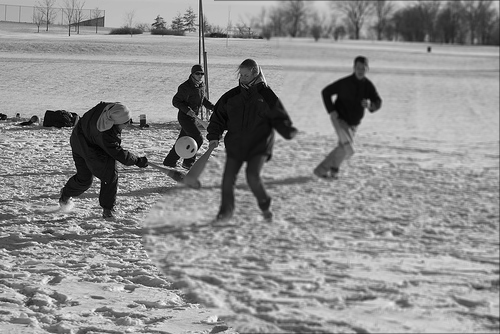}
\caption{The example of source images which have different focus region.}
\label{fig:exampleforsource}
\end{figure}

In our paper, we choose twenty images from ImageNet in sport(http://www.image-net.org/index) as original images. Because the number of original images are too much to show all of them, so we just take an examples for these images, as shown in Fig.\ref{fig:example6}. We blur these original images with Gaussian smoothing filter (size $3$$\times$$3$ and $\sigma=7$) to get twenty pairs images which contain different focus region. And the example of source images shown in Fig.\ref{fig:exampleforsource}.

Secondly, we compare the proposed method with serval typical fusion methods, including: discrete cosine transform fusion method(DWT)\cite{DWT2005}, cross bilateral filter fusion method (CBF)\cite{cbf2013}, discrete cosine harmonic wavelet transform fusion method(DCHWT)\cite{dchwt2013}, sparse-representation-based image fusion approach(SR)\cite{novelsr2016}, sparse representation of classified image patches fusion method(SRCI)\cite{hogsr2017}. In the fusion method SRCI, the training dictionary algorithm is K-SVD and use the Orthogonal Matching Pursuit(OMP) to get sparse coefficients.

%
%

In order to evaluate the fusion performance of fused images which obtained by above fusion methods and the proposed method, three quality metrics are utilized. These are: Average Gradient(AG), Peak Signal to Noise Ratio (PSNR) and Structural Similarity (SSIM). In particular, PSNR and SSIM are reference image based approach and AG is non-reference approach. The fusion performance is better when the increasing numerical index of these three values.

In our experiment, the sliding window size is $8$$\times$$8$, the step is one pixel. The orientation bins of HOG are 6(i.e. $L=6$). The sub-dictionary size is 128 and K-SVD algorithm is used to train sub-dictionaries. The parameter ($\lambda$) of LRR is 100.

%
%

\subsection{Image fusion results} 

In this section, we use twenty pairs source images which contain different focus region to test these contrastive methods and the proposed method. The fused results are shown in Fig.5, we choose one pair source images as an example. And the values of AG, PSNR and SSIM for twenty images are shown in Table 1 and Table 2.

All the experiments are implemented in MTALAB R2016a on 3.2 GHz Intel(R) Core(TM) CPU with 4 GB RAM.

%
%
\begin{figure}[!ht]
\centering
\includegraphics[width=0.3\linewidth]{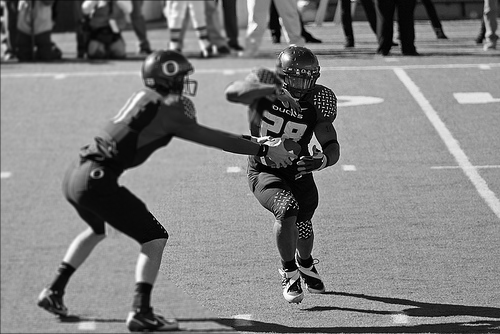}
\includegraphics[width=0.3\linewidth]{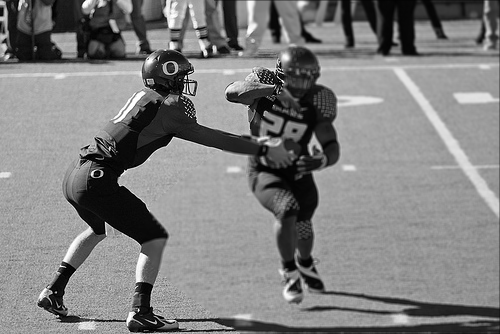}
\includegraphics[width=0.3\linewidth]{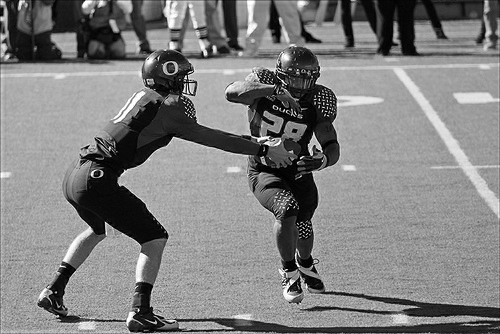} \\
\textnormal{\small(a)source image a} \qquad \quad 
\textnormal{\small(b)source image b} \qquad \qquad \qquad \quad
\textnormal{\small(c)DWT} \\
\includegraphics[width=0.3\linewidth]{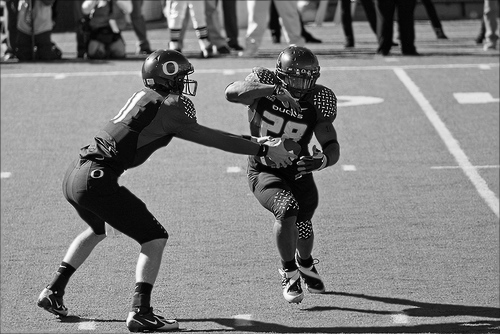}
\includegraphics[width=0.3\linewidth]{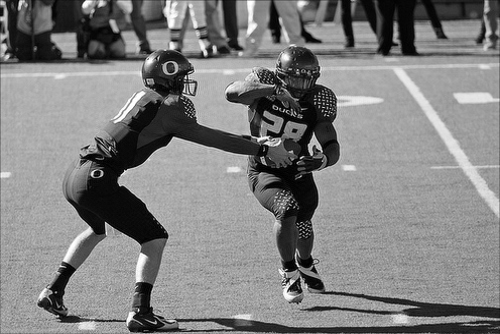}
\includegraphics[width=0.3\linewidth]{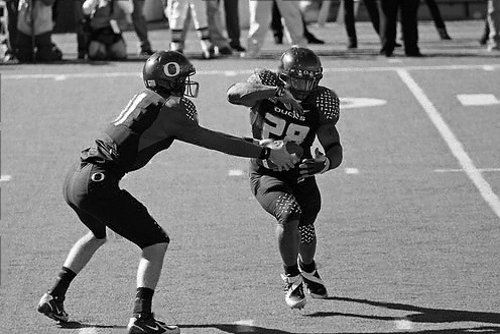} \\
\textnormal{\small(d)CBF} \qquad \qquad \qquad \quad 
\textnormal{\small(e)DCHWT} \qquad \qquad \qquad \quad
\textnormal{\small(f)SR} \\
\includegraphics[width=0.3\linewidth]{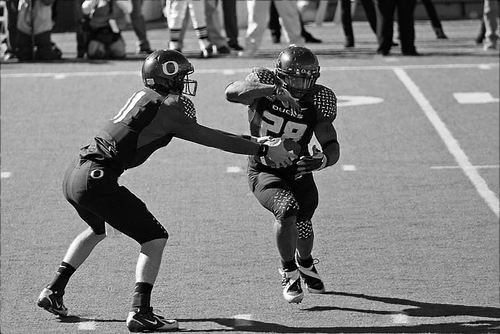}
\includegraphics[width=0.3\linewidth]{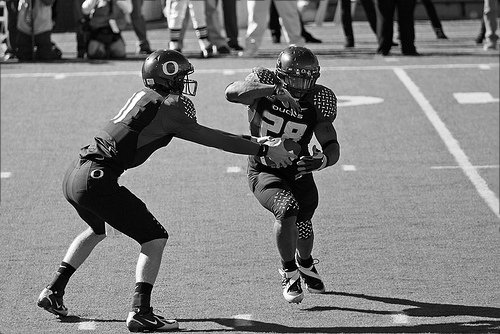} \\
\textnormal{\small(g)SRCI} \qquad \qquad \qquad \quad
\textnormal{\small(h)proposed} \\
\caption{ The example of fuse results. (a) Source image a; (b) Source image b; (c) The fused image obtained by DWT; (d) The fused image obtained by CBF; (e) The fused image obtained by DCHWT; (f) The fused image obtained by SR. (g) The fused image obtained by SRCI; (h) The fused image obtained by the proposed method.}
\label{fig:fuseresults}
\end{figure}

As shown in Fig.\ref{fig:fuseresults}, the fused images obtained by proposed method and other fusion methods are listed. This just shows an example for the fused results which obtained by our experiment. As we can see, the proposed method has same fusion performance compare with other classical and novel fusion methods in human visual system. Therefore we mainly discus the fuse performance with the values of quality metrics, as shown in Table \ref{tab:firsttenimages} and Table \ref{tab:secondtenimages}.

The value of AG indicates the clarity of fused image, the lager value means the fused image is sharper. And the values of PSNR and SSIM denote the difference between fused image and original image. If the fused image has lager values of PSNR and SSIM, then the fused image is more similar to original image.

In Table \ref{tab:firsttenimages} and Table \ref{tab:secondtenimages}, the best results are indicated in bold and the second-best value are indicated in red. As we can see, the proposed fusion method has all best values in AG and nineteen best values and one second-best values in SSIM. And our method has fourteen best values and six second-best values in PSNR. These values indicate that the fused images obtained by the proposed method are more similar to original image and more natural than other fused images obtained by compared methods. 

And the values of SSIM represent the proposed method could preserve the global structure from source images, the values of AG and PSNR denote our fusion method has better performance in local structure. These experiment results demonstrate that the proposed fusion method has better performance than other fusion method.

\section{Conclusions}

In this paper, we proposed a novel fusion method based on dictionary learning and low-rank representation. First of all, the sliding window technique is used to divided the source images. And these image patches are classified by HOG features. Then for each class which obtained by HOG features, we use K-SVD algorithm to train a sub-dictionary. And a global dictionary is obtained by combined these sub-dictionaries. The global dictionary and LRR are used to calculate the low-rank coefficients for each image patches. Then the fused image patches are obtained by $l_1-norm$ and choose-max strategy. Finally, the averaging operation is applied to all overlapping patches to reconstruct the fused image. The experimental results show that the proposed method exhibits better performance than other compared methods.

%
%
\begin{table}[ht]
\centering
\caption{\label{tab:firsttenimages}The AG, PSNR and SSIM values of the compared methods and the proposed method for ten pairs source images.}
\resizebox{\textwidth}{!}{
\begin{tabular}{llllllll}
\hline \noalign{\smallskip}
& & DWT & CBF & DCHWT & SR & SRCI & Proposed \\
\noalign{\smallskip}
\hline
\noalign{\smallskip}
\multirow{3}*{image1} &
AG      & 0.0846      & 0.0748      & 0.0812      & 0.0737      & 0.0800      & \textbf{0.0862} \\
~ & PSNR    & 37.0188     & 34.8557     & \textbf{37.6991}     & 30.8481     & 34.9755     & \textcolor{red}{37.5665} \\
~ & SSIM    & 0.9673      & 0.9647      & 0.9699      & 0.9159      & 0.9581      & \textbf{0.9745} \\
\hline
\noalign{\smallskip}
\multirow{3}*{image2} &
AG      & 0.1049      & 0.0967      & 0.1004      & 0.1004      & 0.0971      & \textbf{0.1068} \\
~ & PSNR    & 36.4171     & 35.5492     & 37.0933     & 32.3211     & 34.1743     & \textbf{40.0973} \\
~ & SSIM    & 0.9783      & 0.9822      & 0.9819      & 0.9461      & 0.9611      & \textbf{0.9949} \\
\hline  
\noalign{\smallskip}
\multirow{3}*{image3} &
AG      & 0.0943      & 0.0842      & 0.0910      & 0.0903      & 0.0886      & \textbf{0.0963} \\
~ & PSNR    & 37.7503     & 35.8778     & 39.4640     & 33.9790     & 35.9429     & \textbf{42.0192} \\
~ & SSIM    & 0.9823      & 0.9824      & 0.9860      & 0.9510      & 0.9750      & \textbf{0.9928} \\
\hline
\noalign{\smallskip}
\multirow{3}*{image4} &
AG      & 0.1028      & 0.0912      & 0.0984      & 0.0948      & 0.0943      & \textbf{0.1041} \\
~ & PSNR    & 36.9723     & 35.0365     & 38.4210     & 31.7981     & 34.8116     & \textbf{41.9717} \\
~ & SSIM    & 0.9795      & 0.9803      & 0.9830      & 0.9185      & 0.9660      & \textbf{0.9918} \\
\hline
\noalign{\smallskip}
\multirow{3}*{image5} &                      
AG      & 0.0907      & 0.0819      & 0.0862      & 0.0884      & 0.0883      & \textbf{0.0924} \\
~ & PSNR    & 37.1535     & 34.9156     & 38.1989     & 33.8048    & 36.1505    & \textbf{40.4725} \\
~ & SSIM    & 0.9785      & 0.9775      & 0.9811      & 0.9554      & 0.9754      & \textbf{0.9891} \\
\hline
\noalign{\smallskip}
\multirow{3}*{image6} &                      
AG      & 0.0847      & 0.0746      & 0.0818      & 0.0802      & 0.0799  & \textbf{0.0863} \\
~ & PSNR    & \textbf{37.0322}     & 33.8555    & 36.7390     & 31.7423     &33.8922    & \textcolor{red}{36.8921} \\
~ & SSIM    & 0.9723      & 0.9707     &  0.9712      & 0.8968      & 0.9617  & \textbf{0.9795} \\
\hline
\noalign{\smallskip}
\multirow{3}*{image7} &                      
AG      & 0.0757      & 0.0674      & 0.0729      & 0.0706      & 0.0714      & \textbf{0.0777} \\
~ & PSNR    & \textbf{37.4557}     & 34.3481     & 36.4605     & 32.7315    & 35.9622    & 36.0487 \\
~ & SSIM    & 0.9741      & 0.9754      & 0.9730      & 0.9217      & 0.9719      & \textbf{0.9817} \\
\hline
\noalign{\smallskip}
\multirow{3}*{image8} &                      
AG      & 0.1708      & 0.1560      & 0.1653      & 0.1632      & 0.1546      & \textbf{0.1739} \\
~ & PSNR    & 34.0008     & 33.7624     & 36.9543     & 29.9209     &30.3544     &\textbf{40.1589} \\
~ & SSIM    & 0.9759      & 0.9794      & 0.9825      & 0.9393      & 0.9490      & \textbf{0.9953} \\
\hline
\noalign{\smallskip}
\multirow{3}*{image9} &                       
AG      & 0.1514      & 0.1378      & 0.1459      & 0.1448      & 0.1393      & \textbf{0.1543} \\
~ & PSNR    & 33.7263     & 33.2206     & 35.3401     & 30.2005    & 31.2392     &\textbf{37.4141} \\
~ & SSIM    & 0.9738      & 0.9784      & 0.9796      & 0.9371      & 0.9534      & \textbf{0.9916} \\
\hline
\noalign{\smallskip}
\multirow{3}*{image10} &                       
AG      & 0.1154      & 0.1030      & 0.1115      & 0.1115      & 0.1092      & \textbf{0.1184} \\
~ & PSNR    & 36.2289     & 34.0351     & 37.7971     & 32.1752    & 34.0357    & \textbf{39.7722} \\
~ & SSIM    & 0.9781      & 0.9788      & 0.9833      & 0.9345      & 0.9672      &\textbf{ 0.9938} \\
\hline
\end{tabular}}
\end{table}

%
%
\begin{table}[ht]
\centering
\caption{\label{tab:secondtenimages}The AG, PSNR and SSIM values of the compared methods and the proposed method for another ten pairs source images.}
\resizebox{\textwidth}{!}{
\begin{tabular}{llllllll}
\hline
& & DWT & CBF & DCHWT & SR & SRCI & Proposed \\
\hline
\noalign{\smallskip}
\multirow{3}*{image11} &
AG      & 0.0680      & 0.0580      & 0.0604      & 0.0605      & 0.0646      & \textbf{0.0702} \\
~ & PSNR    & \textbf{38.3437}     & 34.5043     & 35.8079     & 33.1686     & 35.3299     & \textcolor{red}{36.5102} \\
~ & SSIM    & 0.9750      & 0.9732      & 0.9668      & 0.9332      & 0.9679      & \textbf{0.9779} \\
 \noalign{\smallskip}
 \hline
 \noalign{\smallskip}
\multirow{3}*{image12} &                        
AG      & 0.1142      & 0.1053      & 0.0957      & 0.0958      & 0.1027      & \textbf{0.1159} \\
~ & PSNR    & 35.9972     & 36.9215     & 33.2545     & 32.9198     & 33.6469     & \textbf{42.1478} \\
~ & SSIM    & 0.9793      & 0.9842      & 0.9616      & 0.9340      & 0.9615      & \textbf{0.9929} \\
\hline
\noalign{\smallskip}
\multirow{3}*{image13} &                     
AG      & 0.1109      & 0.0981      & 0.0946      & 0.0949      & 0.1029      & \textbf{0.1133} \\
~ & PSNR    & \textbf{35.2271}     & 32.9351     & 32.9011     & 31.4786     & 33.3810     & \textcolor{red}{34.9402} \\
~ & SSIM    & 0.9694      & 0.9704      & 0.9554      & 0.9158      & 0.9565      & \textbf{0.9742} \\
\hline
\noalign{\smallskip}
\multirow{3}*{image14} &                          
AG      & 0.0853      & 0.0754      & 0.0738      & 0.0748      & 0.0785      & \textbf{0.0866} \\
~ & PSNR    & 36.0482     & 34.5947     & 34.9556     & 33.0427     & 35.3658     & \textbf{39.5008} \\
~ & SSIM    & 0.9770      & 0.9789      & 0.9686      & 0.9237      & 0.9678      & \textbf{0.9919} \\
 \hline
\noalign{\smallskip}
\multirow{3}*{image15} &                         
AG      & 0.1104      & 0.0986      & 0.0977      & 0.0991      & 0.1051      & \textbf{0.1140} \\
~ & PSNR    & \textbf{34.5466}     & 31.5296     & 32.1961     & 29.8237     & 31.4633     & \textcolor{red}{33.0836} \\
~ & SSIM    & 0.9676      & 0.9680      & 0.9602      & 0.9153      & 0.9565      & \textbf{0.9734} \\
 \hline
\noalign{\smallskip}
\multirow{3}*{image16} &                         
AG      & 0.1338      & 0.1181      & 0.1100      & 0.1081      & 0.1204      & \textbf{0.1362} \\
~ & PSNR    & 34.7157     & 33.4927     & 31.4690     & 30.9152     & 32.2771     & \textbf{38.4244} \\
~ & SSIM    & 0.9779      & 0.9803      & 0.9529      & 0.9034      & 0.9535      & \textbf{0.9915} \\
 \hline
\noalign{\smallskip}
\multirow{3}*{image17} &                         
AG      & 0.1947      & 0.1750      & 0.1577      & 0.1530      & 0.1692      & \textbf{0.1975} \\
~ & PSNR    & 33.2829     & 30.9666     & 28.7314     & 29.7126     & 28.9817     & \textbf{36.1011} \\
~ & SSIM    & 0.9748      & 0.9776      & 0.9352      & 0.9286      & 0.9420      & \textbf{0.9886} \\
\hline
\noalign{\smallskip}
\multirow{3}*{image18} &                          
AG      & 0.1721      & 0.1578      & 0.1414      & 0.1361      & 0.1492      &\textbf{0.1748} \\
~ & PSNR    & 32.6045     & 33.4321     & 29.3121     & 29.6468     & 29.3174     & \textbf{37.7719} \\
~ & SSIM    & 0.9715      & 0.9786      & 0.9419      & 0.9321      & 0.9364      & \textbf{0.9901} \\
\hline
\noalign{\smallskip}
\multirow{3}*{image19} &                          
AG      & 0.1425      & 0.1333      & 0.1101      & 0.1046      & 0.1258      & \textbf{0.1437} \\
~ & PSNR    & 36.0427     & 35.9360     & 30.8280     & 32.0196     & 32.2789     & \textbf{37.4533} \\
~ & SSIM    & 0.9792      & 0.9850      & 0.9353      & 0.9324      & 0.9491      & \textbf{0.9873} \\
 \hline
\noalign{\smallskip}
\multirow{3}*{image20} &                         
AG      & 0.0958      & 0.0851      & 0.0800      & 0.0834      & 0.0899      & \textbf{0.0983} \\
~ & PSNR    & 36.1874     & 34.1353     & 33.1515     & 32.8215     & 34.8414     & \textbf{35.9744} \\
~ & SSIM    & 0.9678      & 0.9708      & 0.9586      & 0.9134      & 0.9644      & \textbf{0.9729} \\
 \hline
\end{tabular}}
\end{table}

\end{document}